# Optimal Upper and Lower Bounds for Boolean Expressions by Dissociation

Wolfgang Gatterbauer[*], Dan Suciu

*Computer Science and Engineering, University of Washington, Seattle, WA, USA*

## Abstract

This paper develops upper and lower bounds for the probability of Boolean expressions by treating multiple occurrences of variables as independent and assigning them new individual probabilities. Our technique generalizes and extends the underlying idea of a number of recent approaches which are varyingly called *node splitting*, *variable renaming*, *variable splitting*, or dissociation for probabilistic databases. We prove that the probabilities we assign to new variables are the best possible in some sense.

**Highlights.** • We give upper and lower bounds to the probability of Boolean expressions. • We generalize and extend recent related approaches. • We show the main results in this paper are best possible in some sense.

## 1. Introduction

Several recent papers propose to approximate an intractable counting problem with a tractable relaxed version by treating multiple occurrences of variables, nodes, or tuples as independent, or ignoring some constraints. Choi et al. [1] approximate inference in Bayesian networks by *node splitting*, i.e. removing some dependencies from the original model, and show how the technique subsumes mini-bucket elimination [2]. Ramirez and Geffner [3] treat the problem of obtaining a minimum cost satisfying assignment of a CNF formula by *variable renaming*, i.e. replacing a variable that appears in many clauses by many fresh new variables that appear in few. Pipatsrisawat and Darwiche [4] provide lower bounds for MaxSAT by *variable splitting*, i.e. compiling a relaxation of the original CNF. Andersen et al. [5] relax constraint satisfaction problems by *refinement through node splitting*, i.e. ignoring some interactions between variables. In our recent work [6], we develop a technique called *dissociation* to approximate the ranking of answers to intractable conjunctive queries.

In this paper, we study the probabilities of Boolean expression after treating some occurrences of variables as independent, and assigning them *new individual probability values*. We call this approach *dissociation*. It turns out that the unifying idea of the above papers is to provide either lower bounds for conjunctive expressions, or upper bounds for disjunctive expressions by assigning dissociated variables their original probabilities. We show that these bounds can be understood as duals and give new and non-obvious upper bounds for conjunctive and lower bounds for disjunctive expression (see Fig. 1). We further show that those are optimal in some sense.

We start with some necessary notations (Sect. 2) and definitions (Sect. 3), provide our main results (Sect. 4), treat binary dissociation separately (Sect. 5), illustrate the results with examples (Sect. 6), and give the full proofs in the appendix.

## 2. Notational and Mathematical Background

We use $[k]$ as short notation for $\{1, \ldots, k\}$, write $x_i$ as short notation for $x_i, i \in [k]$ if $k$ is clear from the context or not relevant, use the bar sign for the complement of an event or probability (e.g., $\bar{x} = \neg x$, and $\bar{p} = 1 - p$), and use a bold notation for sets or vectors of variables (e.g., $\mathbf{x} = (x_1, \ldots, x_k)$) alike. Probabilities are always assumed to be between 0 and 1.

Our treatment of Boolean expressions is notably inspired by Crama and Hammer [7], and by Fuhr and Rölleke [8]. We assume a set $\mathbf{x} = \{x_1, \ldots, x_k\}$ of independent Boolean random variables, and assign to each variable $x_i$ a *primitive event* (we do not formally distinguish between the variable $x_i$ and the event $x_i$ that it is true) which is true with probability $\mathbf{P}[x_i] = p_i$. Thus, all primitive events are assumed to be independent (e.g., $\mathbf{P}[x_1 x_2] = p_1 p_2$). We are interested in computing the probabilities of *composed events*, i.e. event expressions $\phi$ which are logically composed of primitive events.

In our formalism, we make further use of a set $\mathbf{A}$ of *complex events*, which are composed from primitive events. It is known that arbitrary correlations can be represented with event expressions starting from independent Boolean random variables only (see Appendix A). The *correlation* $\rho(A, B)$ between Boolean events $A$ and $B$ is defined as $\rho(A, B) = \frac{\text{cov}(A,B)}{\sqrt{\text{var}(A)\text{var}(B)}}$ [9], with covariance $\text{cov}(A, B) = \mathbf{P}[AB] - \mathbf{P}[A]\mathbf{P}[B]$ and variance $\text{var}(A) = \mathbf{P}[A] - (\mathbf{P}[A])^2$. Hence, complex events can be arbitrarily correlated (e.g., $0 \leq \mathbf{P}[AB] \leq \mathbf{P}[A]$, or equally $-1 \leq \rho(A, B) \leq 1$).

We write $\phi(\mathbf{x})$ to indicate that $\mathbf{x}$ is a set of primitive events appearing in the expression $\phi$. Whenever we write $\phi(\mathbf{x}, \mathbf{A})$,

---
[*]Corresponding author
*Email addresses:* gatter@cs.washington.edu (Wolfgang Gatterbauer), suciu@cs.washington.edu (Dan Suciu)

we imply that a set **x** of primitive events and a set **A** of complex events are both appearing in $\phi$, and that **x** are independent of **A**, i.e. the complex events **A** are composed of primitive events different from **x**. For example, $\phi(\mathbf{x}, \mathbf{A})$ may be defined as $\phi(\{x\}, \{A, B\}) = xA \vee xB$, and the complex events $A$ and $B$ as $A := y_1 y_2$ and $B := \bar{y}_1 y_2$ over $\mathbf{y} = (y_1, y_2)$.

The *dual* of a Boolean expression is obtained by exchanging the operators $\vee$ and $\wedge$, as well as the constants 0 and 1 [7, Def. 1.8]. The *duality principle* states that if a Boolean expression is valid, then so is its dual [7, Th. 4.4]. It plays an important role in our paper. The two dual *De Morgan's laws* [7, Th. 1.1.10] state that $\neg(A \vee B) = \bar{A}\bar{B}$ and $\neg(AB) = \bar{A} \vee \bar{B}$. With *absorption*, we refer to the two dual identities [7, Th. 1.1.11]:

$$A \vee B = A\bar{B} \vee B$$
$$AB = (A \vee \bar{B})B .$$

The (disjunctive) *inclusion-exclusion* principle [10, R. 11.8.1] and its less-known conjunctive dual state:

$$\mathbf{P}[A \vee B] = \mathbf{P}[A] + \mathbf{P}[B] - \mathbf{P}[AB] \qquad (1)$$
$$\mathbf{P}[AB] = \mathbf{P}[A] + \mathbf{P}[B] - \mathbf{P}[A \vee B] \qquad (2)$$

From absorption and two special inclusion-exclusion cases:

$$\mathbf{P}[A \vee B] = \mathbf{P}[A] + \mathbf{P}[B] \qquad \text{(if } \mathbf{P}[AB] = 0\text{)}$$
$$\mathbf{P}[AB] = \mathbf{P}[A] + \mathbf{P}[B] - 1 \qquad \text{(if } \mathbf{P}[A \vee B] = 1\text{)}$$

we get the following dual rules, which we call *event splitting*:

$$\mathbf{P}[A \vee B] = \mathbf{P}[A\bar{B}] + \mathbf{P}[B] \qquad (3)$$
$$\mathbf{P}[AB] = \mathbf{P}[A \vee \bar{B}] + \mathbf{P}[B] - 1 . \qquad (4)$$

## 3. Dissociation and Statically-tight Bounds

In this paper, we are interested in statically-tight bounds for dissociated expressions. In this section, we define and illustrate these two concepts. Intuitively, a *dissociation* $\phi'$ of an expression $\phi$ is derived by treating multiple appearances of the same variable as independent, and assigning them individual new probabilities. We are then interested in assigning probabilities to these new variables so that the probability of the dissociated expression $\mathbf{P}[\phi']$ is always either an upper or lower bound for $\mathbf{P}[\phi]$. Furthermore, we want to assign such probabilities which (i) can become tight, and which (ii) guarantee the best bounds possible when ignoring all the other variables. We call such bounds *statically-tight*.

**Definition 3.1** (Dissociation). *A dissociation of a Boolean expression $\phi(\mathbf{x}, \mathbf{A})$ is a new expression $\phi'(\mathbf{x}', \mathbf{A})$ so that there exists a substitution $\theta : \mathbf{x}' \to \mathbf{x}$ that transforms the new into the original expression: $\phi'(\theta(\mathbf{x}', \mathbf{A})) = \phi(\mathbf{x}, \mathbf{A})$. The probability $\mathbf{P}[\phi']$ is evaluated by assigning each new variable $x'_i \in \mathbf{x}'$ independently a new probability $p'_i$.*

**Example 3.2** (Dissociation). *Take the two DNF expressions:*

$$\phi(\mathbf{x}) = \phi(x_1, x_2, x_3, x_4) = x_1 x_3 \vee x_1 x_4 \vee x_2 x_4$$
$$\phi'(\mathbf{x}') = \phi'(x_1, x_2, x_3, x'_4, x''_4) = x_1 x_3 \vee x_1 x'_4 \vee x_2 x''_4 .$$

*Then $\phi'(\mathbf{x}')$ is a dissociation of $\phi(\mathbf{x})$, as $\phi(\mathbf{x}) = \phi'(\theta(\mathbf{x}'))$ for the substitution $\theta = \{(x_1, x_1), (x_2, x_2), (x_3, x_3), (x'_4, x_4), (x''_4, x_4)\}$. Furthermore, assigning $x'_4$ and $x''_4$ the same probability as $x_4$ (i.e. $p''_4 = p'_4 = p_4$) makes $\mathbf{P}[\phi'(\mathbf{x}')]$ an upper bound to $\mathbf{P}[\phi(\mathbf{x})]$. This follows from $\mathbf{P}[\phi(\mathbf{x})] = p_1 p_3 + p_1 p_4 + p_2 p_4 - p_1 p_3 p_4 - p_1 p_3 p_2 p_4 - p_1 p_4 p_2 + p_1 p_3 p_4 p_2$, whereas $\mathbf{P}[\phi'(\mathbf{x}')] = p_1 p_3 + p_1 p_4 + p_2 p_4 - p_1 p_3 p_4 - p_1 p_3 p_2 p_4 - p_1 p_4^2 p_2 + p_1 p_3 p_4^2 p_2$, and thus $\mathbf{P}[\phi'(\mathbf{x}')] - \mathbf{P}[\phi(\mathbf{x})] = (p_1 p_2 p_3 p_4 - p_1 p_2 p_4)(p_4 - 1) \geq 0$. Next consider the two DNF expressions*

$$\psi(\mathbf{y}) = \psi(x_1, x_3, x_4) = x_1 x_3 \vee x_1 x_4 \vee \bar{x}_1 x_4$$
$$\psi'(\mathbf{y}') = \psi'(x_1, x_3, x'_4, x''_4) = x_1 x_3 \vee x_1 x'_4 \vee \bar{x}_1 x''_4 .$$

*Then $\psi'(\mathbf{y}')$ is a dissociation of $\psi(\mathbf{y})$, as $\psi(\mathbf{y}) = \psi'(\theta(\mathbf{y}'))$ for the substitution $\theta = \{(x_1, x_1), (x_3, x_3), (x'_4, x_4), (x''_4, x_4)\}$. Assigning again $p''_4 = p'_4 = p_4$, gives both expressions the same probability $\mathbf{P}[\psi'(\mathbf{y}')] = p_1 p_3 + p_4 - p_1 p_3 p_4 = \mathbf{P}[\psi(\mathbf{y})]$.*

*Both of above dissociations follow a more general template*

$$\omega(\mathbf{z}, \mathbf{A}) = \omega(\{x_4\}, \{A_0, A_1, A_2\}) = A_0 \vee A_1 x_4 \vee A_2 x_4$$
$$\omega'(\mathbf{z}', \mathbf{A}) = \omega'(\{x'_4, x''_4\}, \{A_0, A_1, A_2\}) = A_0 \vee A_1 x'_4 \vee A_2 x''_4$$

*with $A_i$ representing the following composed events: $A_0 = x_1 x_3$ and $A_1 = x_1$ for both, $A_2 = x_2$ for $\phi$, and $A_2 = \bar{x}_1$ for $\psi$. $\omega'(\mathbf{y}')$ is a dissociation of $\omega(\mathbf{y})$, as $\omega(\mathbf{y}) = \omega'(\theta(\mathbf{y}'))$ for the substitution $\theta = \{(x'_4, x_4), (x''_4, x_4)\}$. As we show in this paper, the probability of the dissociation $\mathbf{P}[\omega'(\mathbf{z}', \mathbf{A})]$ is always an upper bound to $\mathbf{P}[\omega(\mathbf{z}, \mathbf{A})]$ irrespective of what expressions are substituted for $\mathbf{A}$, and as long as they are independent of $x_4$, and as long as $p''_4 = p'_4 = p_4$. Also, for some expressions $A_i$, those bounds actually become tight (whenever $A_1$ and $A_2$ are identical). Furthermore, we cannot find values for $p'_4$ and $p''_4$ which give better bounds for all possible $\mathbf{A}$. We call such bounds statically-tight.*

Example 3.2 informally introduces the idea of statically-tight bounds for dissociated expressions. Intuitively, we are interested in bounding the probability of an event expression $\phi(\mathbf{x}, \mathbf{A})$ with another event expression $\psi(\mathbf{x}', \mathbf{A})$, where $\phi$ and $\psi$ use the *same complex events* $\mathbf{A}$ with unknown probabilities and correlations, but *different primitive events* $\mathbf{x}$ and $\mathbf{x}'$ with specified probabilities. In particular, we are interested in "the best" probability assignments $\mathbf{p}'$ to $\mathbf{x}'$ of $\psi$ (i.e. those values that give the tightest bounds) without knowing the probabilities of and correlations between events $\mathbf{A}$. We call such bounds *statically-tight* and define them as follows for the upper case:

**Definition 3.3** (Upper bound). *Given an event expression $\phi(\mathbf{x}, \mathbf{A})$ with $\mathbf{p} = \mathbf{P}[\mathbf{x}]$. Another event expression $\psi(\mathbf{x}', \mathbf{A})$ with $\mathbf{p}' = \mathbf{P}[\mathbf{x}']$ is an upper bound of $\phi$ iff*

$$\forall \mathbf{A} : \mathbf{P}[\phi(\mathbf{x}, \mathbf{A})] \leq \mathbf{P}[\psi(\mathbf{x}', \mathbf{A})] .$$

**Definition 3.4** (Statically-tight upper bound). *An upper bound becomes* statically-tight *iff:*

*(i)*
$$\exists \mathbf{A} : \mathbf{P}[\phi(\mathbf{x}, \mathbf{A})] = \mathbf{P}[\psi(\mathbf{x}', \mathbf{A})] ,$$

*where $\mathbf{A}$ must not be trivial, i.e. $\mathbf{P}[A_i] \neq 0$ and $\mathbf{P}[A_i] \neq 1$.*

*(ii)*
$$\forall \mathbf{p}'' = \mathbf{P}[\mathbf{x}''].\exists \mathbf{A} : \mathbf{P}[\psi(\mathbf{x}', \mathbf{A})] \leq \mathbf{P}[\psi(\mathbf{x}'', \mathbf{A})],$$

where $\mathbf{p}'' \neq \mathbf{p}'$, i.e. there are no other probability assignments for $\mathbf{x}'$ which give tighter bounds.

Lower bounds are defined analogously. We use the name *statically-tight* in order to distinguish from *dynamically-tight* bounds, i.e. assignments of probabilities that are optimal with regard to a particular probability assignment to all variables.[1] Static assignments are made without knowledge of the probabilities of the events $A_i$ nor their mutual correlations. The reason for this restriction is that finding the best approximation for a particular instance may often be NP-hard itself. Restricting oneself to the best possible approximation that can be done without analyzing the dependencies of the non-dissociated variables, i.e. by just considering the syntactic expressions during dissociation, can give very fast algorithms [6].

What we show in this paper is that dissociated expressions together with appropriate probability assignments to the newly dissociated variables can lead to upper *and* lower bounds for the original expressions. Furthermore, we show that there are probability assignments which are statically-tight.

## 4. Statically-tight Bounds for Dissociated Expressions

This section states the main results of this paper.

**Theorem 4.1** (Disjunctive Dissociation). *Let $x$ be a Boolean random variable with probability $p$ and $A_0, \ldots, A_n$ arbitrary Boolean events independent of $x$. Then the probability $\mathbf{P}[\phi_d]$ of the disjunctive expression*

$$\phi_d = A_0 \vee xA_1 \vee xA_2 \vee \ldots \vee xA_n$$

*can be upper and lower bounded by the probability $\mathbf{P}[\phi'_d]$ of its dissociation*

$$\phi'_d = A_0 \vee x_1 A_1 \vee x_2 A_2 \vee \ldots \vee x_n A_n$$

*with $x_i$ as new independent random variables, and*
 (a) $p_i \geq p$ *for upper bounds; and*
 (b) $p_i \leq p$, *s.t.* $\prod_i (1 - p_i) \geq 1 - p$ *for lower bounds.*
*Furthermore, $\mathbf{P}[\phi'_d]$ becomes a statically-tight upper bound for $p_i = p$, and a statically-tight lower bound for $p_i \leq p$, s.t. $\prod_i (1 - p_i) = 1 - p$. Requiring all $p_i$ to be the same, the symmetric statically-tight lower bound results from $p_i = 1 - \sqrt[n]{1 - p}$.*

**Theorem 4.2** (Conjunctive Dissociation). *Let $x$ be a Boolean random variable with probability $p$ and $A_0, \ldots, A_n$ arbitrary*

|  | Disjunctive dissociation | Conjunctive dissociation |
|---|---|---|
| $\mathbf{P}[\phi'] \geq \mathbf{P}[\phi]$ | $p_i = p$ | $p_i = \sqrt[n]{p}$ |
| $\mathbf{P}[\phi'] \leq \mathbf{P}[\phi]$ | $p_i = 1 - \sqrt[n]{1 - p}$ | $p_i = p$ |

Figure 1: Symmetric probability assignments for a disjunctive or a conjunctive dissociation to become a statically-tight upper or lower bound.

*Boolean events independent of $x$. Then the probability $\mathbf{P}[\phi_c]$ of the conjunctive expression*

$$\phi_c = A_0 \wedge (x \vee A_1) \wedge (x \vee A_2) \wedge \ldots \wedge (x \vee A_n)$$

*can be upper and lower bounded by the probability $\mathbf{P}[\phi'_c]$ of its dissociation*

$$\phi'_c = A_0 \wedge (x_1 \vee A_1) \wedge (x_2 \vee A_2) \wedge \ldots \wedge (x_n \vee A_n)$$

*with $x_i$ as new independent random variables, and*
 (a) $p_i \geq p$, *s.t.* $\prod_i p_i \geq p$ *for upper bounds; and*
 (b) $p_i \leq p$ *for lower bounds.*
*Furthermore, $\mathbf{P}[\phi'_c]$ becomes a statically-tight upper bound for $p_i \geq p$, s.t. $\prod_i p_i \geq p$, and a statically-tight lower bound for $p_i = p$. Requiring all $p_i$ to be the same, the symmetric statically-tight upper bound results from $p_i = \sqrt[n]{p}$.*

Figure 1 gives an overview of the symmetric statically-tight bounds for dissociated expressions. Remember that *statically-tight* implies that one cannot find probability assignments for $x_i$ in $\phi'$ that result in generally tighter bounds without knowing the probabilities of and the mutual correlations between $A_i$, and note that *symmetric* implies that all $p_i$ are the same.

## 5. Binary Dissociation Bounds

This section formulates and proves our bounds for the case when a single variable is dissociated into two new independent variables. The general case then simply follows from induction on the number of new variables.

**Proposition 5.1** (Upper Disjunctive Dissociation). *Let $x$ be a random variable with probability $p$, and $A, B, C$ be Boolean events independent of $x$. The probability of*

$$\phi_d = xA \vee xB \vee C$$

*can then be upper bounded by the probability of*

$$\phi^u_d = x_1 A \vee x_2 B \vee C$$

*with $x_i$ as new random variables with $p_i \geq p$. Furthermore, $p_i = p$ is a statically tight upper bound.*

*Proof.* By splitting on event $C$ (Eq. 3), we get

$$\mathbf{P}[\phi_d] = \mathbf{P}[xA\bar{C} \vee xB\bar{C}] + \mathbf{P}[C]$$
$$\mathbf{P}[\phi^u_d] = \mathbf{P}[x_1 A\bar{C} \vee x_2 B\bar{C}] + \mathbf{P}[C].$$

---
[1] Note that this distinction is analogous to *query-centric* vs. *data-centric* rewriting of queries in databases. Query-centric algorithms only consider the syntactic query expression, whereas data-centric algorithms can also make use of particularities in the data instance. The query-centric technique that motivates this paper is *dissociation for probabilistic databases* [6], which approximates the probability scores of queries over probabilistic without looking at the data-instance. Analogously, statically-tight bounds give the best bounds that can be given without evaluating the actual expressions represented by $A_i$.

Hence, the proof follows from comparing the probabilities of

$$\psi_d = xD \vee xE$$
$$\psi_d^u = x_1 D \vee x_2 E$$

with $D := A\bar{C}$ and $E := B\bar{C}$ as new events. From disjunctive inclusion-exclusion (Eq. 1), we get

$$\begin{aligned}\mathbf{P}[\psi_d] &= \mathbf{P}[xD] + \mathbf{P}[xE] - \mathbf{P}[xDE]\\ &= p\mathbf{P}[D] + p\mathbf{P}[E] - p\mathbf{P}[DE]\\ \mathbf{P}[\psi_d^u] &= \mathbf{P}[x_1 D] + \mathbf{P}[x_2 E] - \mathbf{P}[x_1 x_2 DE]\\ &= p_1\mathbf{P}[D] + p_2\mathbf{P}[E] - p_1 p_2\mathbf{P}[DE] \ .\end{aligned}$$

Hence, $\mathbf{P}[\psi_d^u] \geq \mathbf{P}[\psi_d]$, and also $\mathbf{P}[\phi_d^u] \geq \mathbf{P}[\phi_d]$ iff

$$\Delta = (p_1 - p)\mathbf{P}[D] + (p_2 - p)\mathbf{P}[E] + (p - p_1 p_2)\mathbf{P}[DE] \geq 0 \ .$$

This is guaranteed to hold if $p_i \geq p$. From monotonicity of $\Delta$ in $p_1$ and $p_2$ follows that the smallest such bound is given for $p_i = p$. Furthermore, this bound is tight in case that events $A$ and $B$ are disjoint and thus $\mathbf{P}[DE] = \mathbf{P}[AB\bar{C}] = 0$. Since we assume lack of knowledge of the probabilities of and the correlations between $A$ and $B$, the latter bound is statically-tight. □

**Proposition 5.2** (Upper Conjunctive dissociation). *Let $x$ be a random variable with probability $p$, and $A, B, C$ be Boolean events independent of $x$. The probability of*

$$\phi_c = (x \vee A) \wedge (x \vee B) \wedge C$$

*can then be upper bounded by the probability of*

$$\phi_c^u = (x_1 \vee A) \wedge (x_2 \vee B) \wedge C$$

*with $x_i$ as new random variables and choosing $p_i \geq p$, s.t. $p_1 p_2 \geq p$. Furthermore, choosing $p_i \geq p$, s.t. $p_1 p_2 = p$ results in a statically tight upper bound (e.g. by setting $p_i = \sqrt{p}$).*

*Proof.* By splitting on event $C$ (Eq. 4), we get

$$\begin{aligned}\mathbf{P}[\phi_c] &= \mathbf{P}[(x \vee A \vee \bar{C}) \wedge (x \vee B \vee \bar{C})] + \mathbf{P}[C] - 1\\ \mathbf{P}[\phi_c^u] &= \mathbf{P}[(x_1 \vee A \vee \bar{C}) \wedge (x_2 \vee B \vee \bar{C})] + \mathbf{P}[C] - 1 \ .\end{aligned}$$

Hence, the proof follows from comparing the probabilities of

$$\psi_c = (x \vee D) \wedge (x \vee E)$$
$$\psi_c^u = (x_1 \vee D) \wedge (x_2 \vee E)$$

with $D := A \vee \bar{C}$ and $E := B \vee \bar{C}$ as new events. From conjunctive inclusion-exclusion (Eq. 2) and subsequent event splitting (Eq. 3) on $D$, $E$, and $D \vee E$, we get

$$\begin{aligned}\mathbf{P}[\psi_c] &= \mathbf{P}[x \vee D] + \mathbf{P}[x \vee E] - \mathbf{P}[x \vee D \vee E]\\ &= \mathbf{P}[x\bar{D}] + \mathbf{P}[D] + \mathbf{P}[x\bar{E}] + \mathbf{P}[E] - \mathbf{P}[x\bar{D}\bar{E}] - \mathbf{P}[D \vee E]\\ &= p\mathbf{P}[\bar{D}] + p\mathbf{P}[\bar{E}] - p\mathbf{P}[\bar{D}\bar{E}] + \underbrace{\mathbf{P}[D] + \mathbf{P}[E] - \mathbf{P}[D \vee E]}_{\delta :=}\end{aligned}$$

$$\begin{aligned}\mathbf{P}[\psi_c^u] &= \mathbf{P}[x_1 \vee D] + \mathbf{P}[x_2 \vee E] - \mathbf{P}[x_1 \vee x_2 \vee D \vee E]\\ &= \mathbf{P}[x_1\bar{D}] + \mathbf{P}[D] + \mathbf{P}[x_2\bar{E}] + \mathbf{P}[E]\\ &\quad - \mathbf{P}[(x_1 \vee x_2)\bar{D}\bar{E}] - \mathbf{P}[D \vee E]\\ &= p_1\mathbf{P}[\bar{D}] + p_2\mathbf{P}[\bar{E}] - (p_1 + p_2 - p_1 p_2)\mathbf{P}[\bar{D}\bar{E}] + \delta \ .\end{aligned}$$

Hence, $\mathbf{P}[\psi_c^u] \geq \mathbf{P}[\psi_c]$, and also $\mathbf{P}[\phi_c^u] \geq \mathbf{P}[\phi_c]$ iff

$$\Delta = (p_1 - p)(\mathbf{P}[\bar{D}] - \mathbf{P}[\bar{D}\bar{E}]) + (p_2 - p)(\mathbf{P}[\bar{E}] - \mathbf{P}[\bar{D}\bar{E}])$$
$$+ (p_1 p_2 - p)\mathbf{P}[\bar{D}\bar{E}] \geq 0 \ .$$

This is guaranteed to hold if $p_i \geq p$ and $p_1 p_2 \geq p$. Since $\Delta$ is monotone in $p_i$, the smallest such bounds result from setting $p_i \geq p$, s.t. $p_1 p_2 = p$. Furthermore, these bounds are tight if $A$ and $B$ are identical, and hence $\mathbf{P}[\bar{D}] = \mathbf{P}[\bar{E}] = \mathbf{P}[\bar{D}\bar{E}]$. Since we assume lack of knowledge of the probabilities of and correlations between $A$ and $B$, these latter bounds are statically-tight. The symmetric statically-tight bound results from setting $p_1 = p_2$, and hence $p_i = \sqrt{p}$. □

**Proposition 5.3** (Lower Disjunctive Dissociation). *Let $x$ be a random variable with probability $p$, and $A, B, C$ be Boolean events independent of $x$. The probability of*

$$\phi_d = xA \vee xB \vee C$$

*can then be lower bounded by the probability of*

$$\phi_d^l = x_1 A \vee x_2 B \vee C$$

*with $x_i$ as new random variables and choosing $p_i \leq p$, s.t. $(1 - p_1)(1 - p_2) \geq 1 - p$. Furthermore, choosing $p_i \leq p$, s.t. $(1 - p_1)(1 - p_2) = 1 - p$ results in a statically-tight lower bound (e.g. by setting $p_i = 1 - \sqrt{1-p}$).*

*Proof.* Using De Morgan's laws, we can write the complements of $\phi_d$ and $\phi_d^l$ as:

$$\begin{aligned}\bar{\phi}_d &= (\bar{x} \vee \bar{A}) \wedge (\bar{x} \vee \bar{B}) \wedge \bar{C}\\ \bar{\phi}_d^l &= (\bar{x}_1 \vee \bar{A}) \wedge (\bar{x}_2 \vee \bar{B}) \wedge \bar{C}\end{aligned}$$

Treating the complements $\bar{x}, \bar{x}_1, \bar{x}_2, \bar{A}, \bar{B}, \bar{C}, \bar{D} = \bar{A} \vee C$, and $\bar{E} = \bar{B} \vee C$ as new events, and applying Prop. 5.2, we know that $\mathbf{P}[\bar{\phi}_d^l] \geq \mathbf{P}[\bar{\phi}_d]$ if $\bar{p}_i \geq \bar{p}$ and $\bar{p}_1 \bar{p}_2 \geq \bar{p}$. Hence, it follows that $\mathbf{P}[\phi_d^l] \leq \mathbf{P}[\phi_d]$ if $p_i \leq p$ and $(1 - p_1)(1 - p_2) \geq 1 - p$, and that the largest such bounds result from setting $p_i \leq p$, s.t. $(1 - p_1)(1 - p_2) = 1 - p$. From the proof of Prop. 5.2, it further follows that these latter bounds become tight if $A$ and $B$ are identical, and hence $\mathbf{P}[D] = \mathbf{P}[E] = \mathbf{P}[DE]$. Since we assume lack of knowledge of the probabilities of and correlations between $A$ and $B$, these latter bounds are statically-tight. The symmetric statically-tight bound results from setting $p_1 = p_2$, and hence $p_i = 1 - \sqrt{1 - p}$. □

**Proposition 5.4** (Lower Conjunctive Dissociation). *Let $x$ be a random variable with probability $p$, and $A, B, C$ be Boolean events independent of $x$. The probability of*

$$\phi_c = (x \vee A) \wedge (x \vee B) \wedge C$$

*can then be lower bounded by the probability of*

$$\phi_c^l = (x_1 \vee A) \wedge (x_2 \vee B) \wedge C$$

*with $x_i$ as new random variables with $p_i \leq p$. Furthermore, $p_i = p$ is a statically tight lower bound.*

*Proof.* Using De Morgan's laws, we can write the complements of $\phi_c$ and $\phi_c^l$ as:

$$\bar{\phi}_c = \bar{x}\bar{A} \vee \bar{x}\bar{B} \vee \bar{C}$$

$$\bar{\phi}_c^l = \bar{x}_1\bar{A} \vee \bar{x}_2\bar{B} \vee \bar{C}$$

Treating the complements $\bar{x}, \bar{x}_1, \bar{x}_2, \bar{A}, \bar{B}, \bar{C}, \bar{D} = \bar{A}C$, and $\bar{E} = \bar{B}C$ as new events, and applying Prop. 5.1, we know that $\mathbf{P}[\bar{\phi}_c^l] \geq \mathbf{P}[\bar{\phi}_c]$ if $\bar{p}_i \geq \bar{p}$. Hence, it follows that $\mathbf{P}[\phi_c^l] \leq \mathbf{P}[\phi_c]$ if $p_i \leq p$, and that the largest such bound results from $p_i = p$. From the proof of Prop. 5.1, it further follows that this bound is tight if $\bar{A}$ and $\bar{B}$ are disjoint (i.e. $\mathbf{P}[A \vee B = 1]$) and hence $\mathbf{P}[\bar{D}\bar{E}] = \mathbf{P}[\bar{A}\bar{B}C] = 0$. Since we assume lack of knowledge of the probabilities of and correlations between $D$ and $E$, this bound is statically-tight. □

## 6. Illustration

We illustrate our symmetric statically-tight bounds for the disjunctive expression $\phi_d = xA \vee xB$ and the conjunctive expression $\phi_c = (x \vee A)(x \vee B)$. Let $p = \mathbf{P}[x]$, $q = \mathbf{P}[A] = \mathbf{P}[B]$, and assume $x$ to be independent of $A$ and $B$, which can be arbitrarily correlated: $-1 \leq \rho(A, B) \leq 1$. Further, let $p_i = \mathbf{P}[x_1] = \mathbf{P}[x_2]$ be the probabilities in the dissociated expressions.

In a few steps, one can calculate the probabilities of $\phi_d$, $\phi_c$ and their dissociations as

$$\mathbf{P}[\phi_d] = 2pq - p\mathbf{P}[AB]$$
$$\mathbf{P}[\phi_d'] = 2p_iq - p_i^2\mathbf{P}[AB]$$
$$\mathbf{P}[\phi_c] = p + (1-p)\mathbf{P}[AB]$$
$$\mathbf{P}[\phi_c'] = 2p_iq + p_i^2(1-2q) + (1-p_i^2)\mathbf{P}[AB]$$

Figure 2 illustrates the probabilities of the expressions and their symmetric statically-tight upper and lower dissociations for various values of $p$, $q$ and as function of the correlation $\rho(A, B)$, and by setting $p_i$ according to Fig. 1. Remember that $\rho(A, B) = \frac{\mathbf{P}[AB]-q^2}{q-q^2}$ and, hence: $\mathbf{P}[AB] = \rho(A, B) \cdot (q - q^2) + q^2$. Further, $\mathbf{P}[AB] = 0$ (i.e. disjointness between $A$ and $B$) is not possible for $q > 0.5$, and from $\mathbf{P}[A \vee B] \leq 1$, one can derive $\mathbf{P}[AB] \geq 2p-1$. In turn, $\rho = -1$ is not possible for $q < 0.5$, and it must hold $\mathbf{P}[AB] \geq 0$. From both together, one can derive the condition $\rho_{\min}(q) = \max(-\frac{q}{1-q}, -\frac{1+q^2-2q}{q-q^2})$. This marks the beginning of the graphs for different values of $q$ in the figure.

## Acknowledgements

We like to thank Adnan Darwiche for helpful discussions. This work was partially supported by NSF grant IIS-0915054, and was performed in the context of the LaPushDB project (http://LaPushDB.com/).

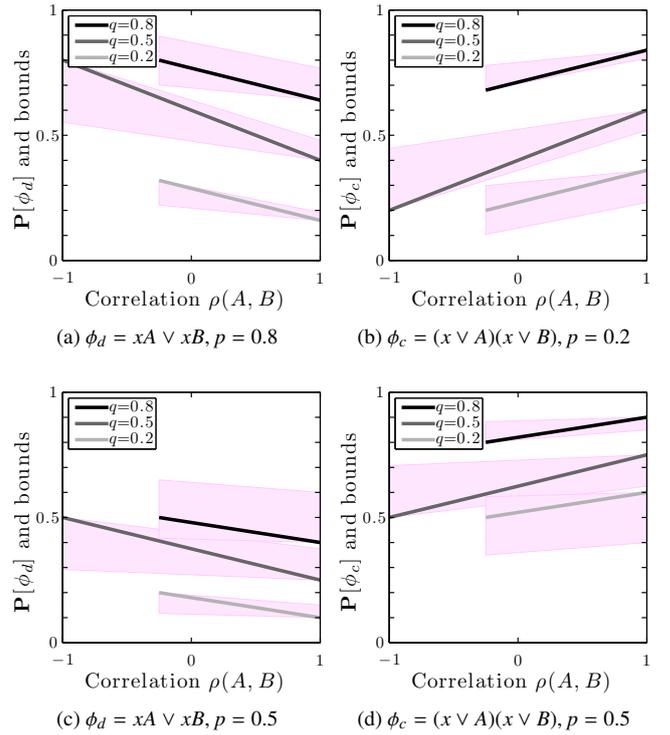

Figure 2: Probabilities of disjunctive and conjunctive expression together with their symmetric statically-tight bounds from dissociation as function of the correlation between $A$ and $B$. We use $p = \mathbf{P}[x]$ and $q = \mathbf{P}[A] = \mathbf{P}[B]$.

## Appendix A. Representing Complex Events

It is known from Poole's independent choice logic [11] that arbitrary correlations between events can be composed from *disjoint-independent events* only. A disjoint-independent event is represented by a non-Boolean independent random variable $x$ which takes either of $k$ values $\mathbf{v} = (v_1, \ldots, v_k)$ with respective probabilities $\mathbf{p} = (p_1, \ldots, p_k)$ and $\sum_i p_i = 1$. Poole writes such a "disjoint declaration" as $x([v_1 : p_1, \ldots, v_k : p_k])$.

In turn, any $k$ disjoint events can be represented starting from $k - 1$ independent Boolean variables $\mathbf{y} = (y_1, \ldots, y_{k-1})$ and probabilities $\mathbf{P}[\mathbf{y}] = (p_1, \frac{p_2}{\bar{p}_1}, \frac{p_3}{\bar{p}_1 \bar{p}_2}, \ldots, \frac{p_{k-1}}{\bar{p}_1 \cdots \bar{p}_{k-2}})$, by assigning the disjoint-independent event variable $x$ its value $v_i$ whenever event $A_i$ is true with $A_i$ defined as:

$$(x = v_1) \equiv A_1 := x_1$$
$$(x = v_2) \equiv A_2 := \bar{x}_1 x_2$$
$$\vdots$$
$$(x = v_{k-1}) \equiv A_{k_1} := x_1 \ldots \bar{x}_{k-2} x_{k-1}$$
$$(x = v_k) \equiv A_k := \bar{x}_1 \ldots \bar{x}_{k-2} \bar{x}_{k-1} .$$

For example, a primitive disjoint-independent event variable $x(v_1 : \frac{1}{5}, v_1 : \frac{1}{2}, v_1 : \frac{1}{5}, v_1 : \frac{1}{10})$ can be represented with three independent Boolean variables $\mathbf{y} = (y_1, y_2, y_3)$ and $\mathbf{P}[\mathbf{y}] = (\frac{1}{5}, \frac{5}{8}, \frac{2}{3})$.

It follows that *arbitrary correlations between events* can be modeled starting from *independent Boolean random variables* alone. For example, two *complex events* $A$ and $B$ with $\mathbf{P}[A] = \mathbf{P}[B] = q$ and varying correlation (see Sect. 6) can be represented as *composed events* $A := y_1 y_2 \lor y_3 \lor y_4$ and $B := \bar{y}_1 y_2 \lor y_3 \lor y_5$ over the *primitive events* $\mathbf{y}$ with varying probabilities $\mathbf{P}[\mathbf{y}]$. Events $A$ and $B$ become identical for $\mathbf{P}[\mathbf{y}] = (0, 0, q, 0, 0)$, independent for $\mathbf{P}[\mathbf{y}] = (0, 0, 0, q, q)$, and disjoint for $\mathbf{P}[\mathbf{y}] = (0.5, q, 0, 0, 0)$ with $q \leq 0.5$.

## Appendix B. Proofs of Main Theorems

*Proof Theorem 4.1 (a).* The proof follows from Prop. 5.1 and induction on the number of new variables for each original dissociated variable. Assume we know that $\mathbf{P}[\phi_d^k] \geq \mathbf{P}[\phi_d]$ for

$$\phi_d = A_0 \lor x A_1 \lor \ldots \lor x A_n = A_0 \lor x(A_1 \lor \cdots \lor A_n)$$
$$\phi_d^k = A_0 \lor x_1 A_1 \lor \cdots \lor x_{k-1} A_{k-1} \lor x_k (A_k \lor \cdots \lor A_n)$$

with $p_i \geq p, i \in [k]$ and $k < n$. We need to show that then $\mathbf{P}[\phi_c^{k+1}] \geq \mathbf{P}[\phi_c]$, with $p_i \geq p, i \in [k+1]$ for

$$\phi_d^{k+1} = A_0 \lor x_1 A_1 \lor \cdots \lor x_k A_k \lor x_{k+1}(A_{k+1} \lor \cdots \lor A_n) .$$

Applying Prop. 5.1 to $\phi_d^k$, we know $\mathbf{P}[\phi_d^{k'}] \geq \mathbf{P}[\phi_d^k]$ for

$$\phi_d^{k'} = A_0 \lor x_1 A_1 \lor \cdots \lor x_{k-1} A_{k-1} \lor x_k' A_k \lor x_k''(A_{k+1} \lor \cdots \lor A_n)$$

with $p_k' \geq p_k \geq p$ and $p_k'' \geq p_k \geq p$. Hence, we have shown $\mathbf{P}[\phi_d^{k+1}] \geq \mathbf{P}[\phi_d^k] \geq \mathbf{P}[\phi_d]$ after substituting $p_k \leftarrow p_k'$ and $p_{k+1} \leftarrow p_k''$. Since $\phi_d'$ is monotone in $p_i$, it follows that the smallest such upper bound results from choosing $p_i = p$. Furthermore, this bound is tight in case the events $A_i, i \in [n]$ are disjoint, since then $\mathbf{P}[\phi_d^u] = \mathbf{P}[A_0] + p\mathbf{P}[A_1 \bar{A}_0] + \cdots + p\mathbf{P}[A_n \bar{A}_0] = \mathbf{P}[\phi_d]$. Since we assume lack of knowledge of the probabilities of and correlations between events $A_i$, the bound is statically tight. $\square$

*Proof Theorem 4.2 (a).* The proof follows from Prop. 5.2 and induction on the number of new variables for each original dissociated variable. Assume we know that $\mathbf{P}[\phi_c^k] \geq \mathbf{P}[\phi_c]$ for

$$\phi_c = A_0 \land (x \lor A_1) \land \cdots \land (x \lor A_n) = A_0 \land (x \lor A_1 \cdots A_n)$$
$$\phi_c^k = A_0 \land (x_1 \lor A_1) \land \cdots \land (x_{k-1} \lor A_{k-1}) \land (x_k \lor A_k \cdots A_n)$$

with $p_i \geq p$, s.t. $\prod_{i \in [k]} p_i \geq p$, and $k < n$. We need to show that then $\mathbf{P}[\phi_c^{k+1}] \geq \mathbf{P}[\phi_c]$, with $p_i \geq p$, s.t. $\prod_{i \in [k+1]} p_i \geq p$ for

$$\phi_c^{k+1} = A_0 \land (x_1 \lor A_1) \land \cdots \land (x_k \lor A_k) \land (x_{k+1} \lor A_{k+1} \cdots A_n) .$$

Applying Prop. 5.2 to $\phi_c^k$, we know $\mathbf{P}[\phi_c^{k'}] \geq \mathbf{P}[\phi_c^k]$ for

$$\phi_c^{k'} = A_0 \land (x_1 \lor A_1) \land \cdots \land (x_{k-1} \lor A_{k-1}) \land (x_k' \lor A_k) \land (x_k'' \lor A_{k+1} \cdots A_n)$$

with $p_k' \geq p_k \geq p$, $p_k'' \geq p_k \geq p$, and $p_k' p_k'' \geq p_k \geq p$. Hence, it follows that $p_1 \cdots p_{k-1} p_k' p_k'' \geq p_1 \cdots p_k \geq p$. Hence, we have shown $\mathbf{P}[\phi_c^{k+1}] \geq \mathbf{P}[\phi_c^k] \geq \mathbf{P}[\phi_c]$ after substituting $p_k \leftarrow p_k'$ and $p_{k+1} \leftarrow p_k''$. Since $\phi_c'$ is monotone in $p_i$, it follows that the smallest such upper bounds result from choosing $p_i \geq p$, s.t. $\prod_{i \in [n]} p_i = p$. Furthermore, these bounds are tight in case $A_i, i \in [n]$ are identical, since then $\mathbf{P}[\phi_c'] = \mathbf{P}[(x_1 \cdots x_n)(A_1 \lor \bar{A}_0)] + \mathbf{P}[A_0] - 1 = p\mathbf{P}[A_1 \lor \bar{A}_0] + \mathbf{P}[A_0] - 1 = \mathbf{P}[\phi_c]$. Since we assume lack of knowledge of the probabilities of and correlations between events $A_i$, these latter bounds are statically-tight. The symmetric statically-tight bound results from additionally requiring $p_i = p_j$, and thus setting $p_i = \sqrt[n]{p}$. $\square$

*Proof Theorem 4.1 (b).* Using De Morgan's laws, we can write the complements of $\phi_d$ and $\phi_d'$ as:

$$\bar{\phi}_d = \bar{A}_0 \land (\bar{x} \lor \bar{A}_1) \land \cdots \land (\bar{x} \lor \bar{A}_n)$$
$$\bar{\phi}_d' = \bar{A}_0 \land (\bar{x}_1 \lor \bar{A}_1) \land \cdots \land (\bar{x}_n \lor \bar{A}_n)$$

From Theorem 4.2 (a) we know that $\mathbf{P}[\bar{\phi}_d'] \geq \mathbf{P}[\bar{\phi}_d]$ if $\bar{p}_i \geq \bar{p}$, s.t. $\prod_i \bar{p}_i \geq \bar{p}$. Hence, it follows that $\mathbf{P}[\phi_d'] \leq \mathbf{P}[\phi_d]$ if $p_i \leq p$, s.t. $\prod_i (1 - p_i) \geq 1 - p$, and that the smallest such bounds result from $p_i \leq p$, s.t. $\prod_i (1 - p_i) = 1 - p$. Furthermore, these later bounds are tight in case $A_i, i \in [n]$ are identical, since then $\mathbf{P}[\phi_d'] = \mathbf{P}[(x_1 \lor \cdots \lor x_n) A_1 \bar{A}_0] + \mathbf{P}[A_0] - 1 = p\mathbf{P}[A_1 \bar{A}_0] + \mathbf{P}[A_0] - 1 = \mathbf{P}[\phi_d]$. Since we assume lack of knowledge of the probabilities of and correlations between events $A_i$, these latter bounds are statically-tight. The symmetric statically-tight lower bound results from additionally requiring $p_i = p_j$, and thus setting $p_i = 1 - \sqrt[n]{1 - p}$. $\square$

*Proof Theorem 4.2 (b).* Using De Morgan's laws, we can write the complements of $\phi_c$ and $\phi_c'$ as:

$$\bar{\phi}_c = \bar{A}_0 \lor \bar{x} \bar{A}_1 \lor \ldots \lor \bar{x} \bar{A}_n$$
$$\bar{\phi}_c' = \bar{A}_0 \lor \bar{x}_1 \bar{A}_1 \lor \ldots \lor \bar{x}_n \bar{A}_n$$

From Theorem 4.1 (a) we know that $\mathbf{P}[\bar{\phi}_c'] \geq \mathbf{P}[\bar{\phi}_c]$ if $\bar{p}_i \geq \bar{p}$. Hence, it follows that $\mathbf{P}[\phi_c'] \leq \mathbf{P}[\phi_c]$ if $p_i \leq p$, and that the smallest such bound results from setting $p_i = \bar{p}$. Furthermore this latter bound is tight in case $\bar{A}_i, i \in [n]$ are disjoint (this is true if $\mathbf{P}[A_i \lor A_j] = 1$), since then $\mathbf{P}[\bar{\phi}_c'] = \mathbf{P}[\bar{A}_0] + (1-p)\mathbf{P}[\bar{A}_1] + \cdots + (1-p)\mathbf{P}[\bar{A}_n] = \mathbf{P}[\bar{\phi}_c]$. Since we assume lack of knowledge of the probabilities of and correlations between events $A_i$, this bound is statically-tight. $\square$

# Appendix C. Nomenclature

| | |
|---|---|
| $x, y, z$ | primitive events: independent Boolean random variables |
| $\phi, \psi, \omega$ | composed events: Boolean event expressions |
| $A, B, \ldots$ | complex events: events composed from primitive events with unknown event expressions |
| $\mathbf{P}[A], \mathbf{P}[\phi]$ | probability of an event or expression |
| $p_i$ | probability $\mathbf{P}[x_i]$ of random variable $x_i$ |
| $\mathbf{x}, \mathbf{A}, \mathbf{p}$ | ordered sets $(x_1, \ldots, x_k)$ or unordered sets $\{x_1, \ldots, x_k\}$ of variables, events or probabilities |
| $\bar{x}, \bar{A}, \bar{\phi}, \bar{p}$ | complements: $\neg x, \neg A, \neg \phi, 1 - p$ |